\documentclass{article}
\usepackage{spconf,amsmath,epsfig,booktabs,subfig,color,multirow,balance,amssymb,comment,url}

\let\OLDthebibliography\thebibliography
\renewcommand\thebibliography[1]{
  \OLDthebibliography{#1}
  \setlength{\parskip}{0pt}
  \setlength{\itemsep}{0pt plus 0.3ex}
}

\begin{document}\sloppy

% Example definitions.
% --------------------
\def\x{{\mathbf x}}
\def\L{{\cal L}}

% Title.
% ------
\title{Towards Coding for Human and Machine Vision: \\A Scalable Image Coding Approach}
%
% Single address.
% ---------------
\name{Yueyu Hu, Shuai Yang, Wenhan Yang, Ling-Yu Duan and Jiaying Liu\vspace{-4mm}}
%Address and e-mail should NOT be added in the submission paper. They should be present only in the camera ready paper.
\address{Peking University, Beijing, China}

\maketitle

\begin{abstract}
The past decades have witnessed the rapid development of image and video coding techniques in the era of big data. However, the signal fidelity-driven coding pipeline design limits the capability of the existing image/video coding frameworks to fulfill the needs of both machine and human vision. In this paper, we come up with a novel image coding framework by leveraging both the compressive and the generative models, to support machine vision and human perception tasks jointly. Given an input image, the feature analysis is first applied, and then the generative model is employed to perform image reconstruction with features and additional reference pixels, in which compact edge maps are extracted in this work to connect both kinds of vision in a scalable way. The compact edge map serves as the basic layer for machine vision tasks, and the reference pixels act as a sort of enhanced layer to guarantee signal fidelity for human vision. By introducing advanced generative models, we train a flexible network to reconstruct images from compact feature representations and the reference pixels. Experimental results demonstrate the superiority of our framework in both human visual quality and facial landmark detection, which provide useful evidence on the emerging standardization efforts on MPEG VCM (Video Coding for Machine)\footnote{https://lists.aau.at/mailman/listinfo/mpeg-vcm}. Our project website is available at \url{https://williamyang1991.github.io/projects/VCM-Face/}.
\end{abstract}
\vspace{-3mm}
\section{Introduction}
\label{sec:intro}
\vspace{-2mm}
% HYY. MVC prospective.
Image compression has been one of the most fundamental techniques in media sharing and storage.
The typical goal of image compression is to preserve as much signal fidelity with the bit-rate constraint.
The mainstream hybrid coding scheme for images, such as JPEG and JPEG 2000 which typically include transform, quantization, and entropy coding modules, has been developed for decades. It improves the signal fidelity-driven metrics significantly and benefits human vision continuously.

However, in the big data era when massive amounts of data generated everyday needs to be compressed, stored and analyzed, existing compression methods get into troubles to fulfill the needs of both machine and human vision. The sequential compression and analysis paradigm is expensive and even intractable if we expect to maintain the quality of the reconstructed videos. On the other way, when the compression ratio is high~\cite{rdo_feat}, the performance of machine vision tasks degrades significantly.

%Previous human vision-driven compression methods run into problems
%when encountering big data and video analytics.
%The massive data streaming generated everyday from the smart cities
%needs to be compressed, transmitted and analyzed to provide high valuable  information, such as the results of action recognition, event detection, \textit{etc}.
%Given this scenario, it is expensive to perform the analysis on the compressed videos,
%as the video coding bit-stream is redundant and existing coding mechanism
%is not flexible to first discard the information that is not related to analytical tasks~\cite{rdo_feat}.
%Therefore, in the context of big data, it is still an open problem to perform the scalable video coding,
%where the requirement of machine vision is first met and additional bit-rates can be utilized to further improve visual quality of the reconstructed video progressively and incrementally.
%It is an urgent need to obtain a scalable feature representation that connects the information of low and high-level visions
%and switch the forms between two purposes freely.

%The widely used rate-distortion metrics are assumed to reflect how good the visual quality can be with the bit-rate constraints.
Several works have made efforts in addressing the problem of video analytics on massive data by directly extracting and compressing features used for machine vision tasks into a compact form, rather than compressing the whole high-quality videos.
Several typical features are developed, \textit{e.g.} Scale-Invariant Feature Transform~(SIFT)~\cite{lowe2004distinctive}, Compact descriptors for visual search~(CDVS)~\cite{CDVS} for image understanding, and skeleton for human action recognition~\cite{song2017end}. In this way, the process of feature extraction, compression and transmission becomes light-weighted and less amount of bitstreams are to be handled.

%With the booming of machine intelligence applications, there is an emerging demand for image coding techniques to facilitate the development of the machine vision systems.
%Different from coding for human perception where information in the source image should be at most preserved, it is enough to drive machine vision systems with only a very compact and sparse feature representation of the original image or video, \textit{e.g.} SIFT~\cite{lowe2004distinctive} for image understanding and skeleton for human action recognition~\cite{song2017end}. It takes much less bit-rate to store and transmit such feature representation.

Though these features are compact and highly effective for machine vision tasks, they cannot support machine and human vision tasks jointly in a flexible way, which is expected in the new coding paradigm of video coding for machine~(VCM). This is due to the huge gap between feature coding for machine vision and signal encoding for human vision.
Existing solutions only pay attention to one of these two aspects.
In the big data context, it is still an open problem to support a scalable coding paradigm to satisfy both kinds of vision.
Some works show potential ways to address this problem.
In~\cite{chang2019layered,gregor2016towards}, generative models are used to reconstruct the images based on the encoded features with very few bits towards conceptual coding.
In~\cite{torfason2018towards}, the bitstreams generated by a Variational Auto-Encoder (VAE) is used for image understanding.
However, these attempts are still far away from the ideal targets of VCM:
the requirement of machine vision is first satisfied to provide the fast analysis,
and more bit-rates are additionally used to further improve the visual quality in the reconstruction.

%are to specially design either for machine consumption or human perception, where two separate codecs will be needed, making the system less scalable in terms of human and machine consumption. There is a need for a scalable image coding framework to encode informative image features that can be jointly used for machine vision and image  reconstruction.

%Earlier approaches to feature based compression~\cite{chang2019layered,gregor2016towards} seek to extract compact features to represent images and employ generative models to reconstruct the images based on the encoded features.
%They show the potential of generative models to recover images from very little information, thus facilitating very low-bit-rate compression of images. However, the frameworks are designed only for human vision and whether such feature representations can be utilized for machine vision tasks is unclear.
%In~\cite{torfason2018towards}, the bitstreams generated by a Variational Auto-Encoder (VAE) based image codec are directly used to conduct image understanding without decoding. Though the feature can be utilized by both machine and human vision, it is non-scalable as the full bit-rate representation is required to conduct machine vision analysis.

In our work, we take a further step to bridge the gap between image compression for both machine and human vision.
By leveraging both compressive and generative models, a scalable image coding framework is constructed to support machine and human vision tasks jointly. In this framework, the source image is represented via a compressive model as edge maps and sparse key reference pixels. The edges are parameterized into vectors as the base layer of the coding bits to obtain a compact feature representation, which only takes a small portion of coding bits.
Furthermore, the information in our edge maps is shown to be efficient for machine vision tasks, \textit{e.g.} face landmark detection.
To better reconstruct the high-quality frame, reference pixels, sampled in accordance with the edges, can be transmitted as a second layer to the decoder.
With the reference pixel values, the decoder is able to faithfully reconstruct the image.
We adopt a generative model to reconstruct high-quality images from the sparse edge representations.
Experiments on both machine and human vision show significant improvements compared with existing methods, which provide useful evidence on the emerging standardization efforts on MPEG VCM.

In summary, the contributions of this work are threefold:\vspace{-3mm}
\begin{itemize}
\setlength{\itemsep}{0pt}
\setlength{\parsep}{0pt}
\setlength{\parskip}{0pt}
  \item We propose an image coding framework that leverages the compressive model to extract highly compact representations of an image and faithfully reconstruct the original image from the bitstreams with the generative model.
  \item We design the vision-driven compact representations for image compression, where the critical image structure and color information is sparsely encoded. A deep generative network is further proposed to effectively recover images from our compact representations.
  \item A good balance between human and machine vision is stricken, where we achieve $90\%$ and $73\%$ human vision preferences in terms of fidelity and aesthetics, respectively, and achieve an error drop of $44.75\%$ in the machine vision facial landmark detection task.\vspace{-3mm}
\end{itemize}

The rest of this paper is organized as follows.~Section~\ref{sec:relatedwork} reviews related works.~Section~\ref{sec:alg} presents the proposed scalable image coding method.~Experimental results are shown in Section~\ref{sec:experiment} and concluding remarks are given in Section~\ref{sec:conclusion}.

% \vspace{-2mm}
\section{Related Work}
\label{sec:relatedwork}
\vspace{-1mm}

\textbf{Feature based Image Coding}. Besides the mainstream transform based codecs~\cite{wallace1992jpeg,christopoulos2000jpeg2000}, there have been other approaches to explore encoding representative image features for reconstruction. In \cite{gregor2016towards}, a generative compression framework is proposed to encode an image into low-bit-rate latent code and exploit recurrent generative networks for reconstruction. With compressive variational auto-encoders (VAE)~\cite{theis2017lossy}, generative networks are also utilized in \cite{chang2019layered} to reconstruct images from edges and latent features produced by neural networks. Though these frameworks encode compact feature representations of images, they are not shown to both satisfy the need of human and machine vision. In \cite{torfason2018towards}, a deep-based encoder is designed to produce latent code that simultaneously serves for machine vision tasks and image reconstruction. However, the encoded feature representation is non-scalable as the full bit-stream is needed to support the machine vision tasks, neglecting the sparsity features for machine vision. In this work, we explore to encode a base layer of features to facilitate machine vision and an additional layer to improve signal fidelity.

\noindent\textbf{Image Generation}.~Image generation aims to generate new images.
Recent image generation methods focus on the powerful generative adversarial networks (GAN)~\cite{Goodfellow2014Generative} to learn data distribution using two adversarial networks. By incorporating additional information such as the text, labels, segmentation maps and edges as inputs, users are able to control the output with these conditions. The advanced GAN has shown impressive capability of data distribution learning to recover abundant information that well matches human visions from limited conditions. Such an advantage is also verified by the closely related image inpainting task, where plausible image content is generated from very sparse contextual information~\cite{dekel2018sparse}. It demonstrates the potential for vision-driven image coding, which forms our research focus in this paper.

\begin{figure*}[t]
  \centering
  \includegraphics[width=0.98\linewidth]{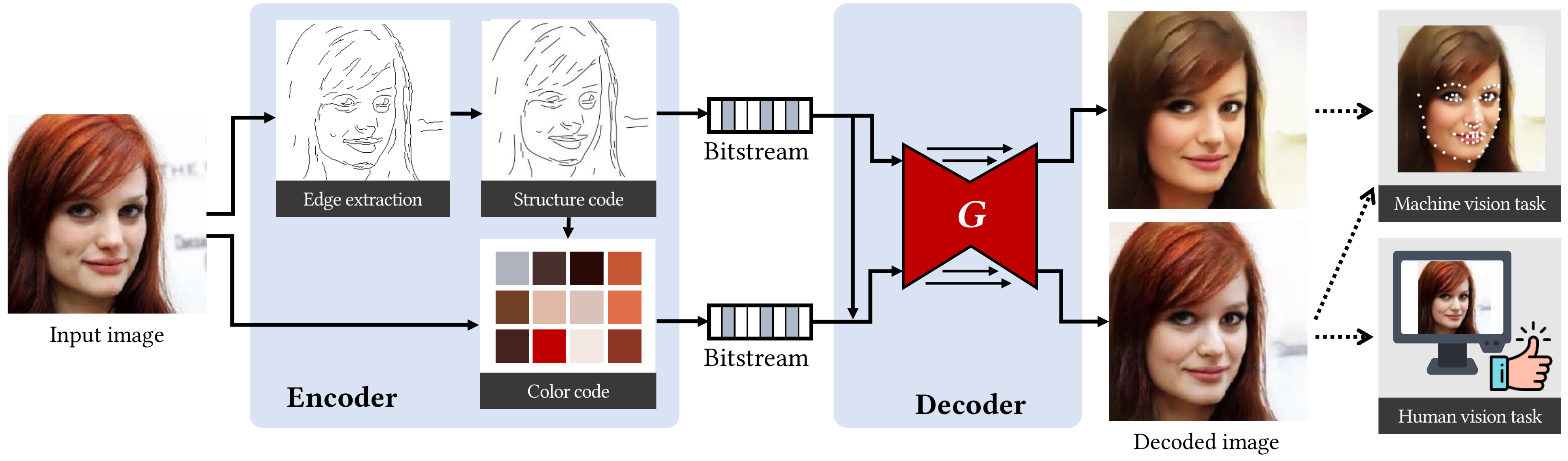}%\vspace{-3mm}
  \caption{Overview of the proposed vision-driven image coding framework.}
  \label{fig:overview}\vspace{-3mm}
\end{figure*}

\section{Proposed Method}
\label{sec:alg}

In this section, we describe our vision-driven image compression framework.
As shown in Fig.~\ref{fig:overview}, we first extract sparse edges to depict the key structure information of the input image~(Section~\ref{sec:extract}).
We then extract vision-driven compact representations through compressive analysis over the edges and the original images.~(Section~\ref{sec:compress}).
Finally, we train a deep neural network to reconstruct the original image from our compact representation.~(Section~\ref{sec:reconstruct}).

\subsection{Sparse Edge Extraction}
\label{sec:extract}

Edges are one of the most highly abstract and sparse image representations.
Edges depict the key structure information of the image, which is consistent with the human vision.
Humans are able to identify the objects from several lines and even infer fine details such as the colors and textures.
To this end, we are inspired to build our compact representation using sparse edges. We will show later that images can be plausibly reconstructed purely from its edges based on the robust data distribution learned by GAN.

Specifically, for an input image $I$, we first use fast edge detection~\cite{DollarICCV13edges} based on structured forests to detect the edge map of $I$. Then, we follow the post process suggested by pix2pix~\cite{Isola2017Image} to binarize the edge maps and discard trivial edges that contain less than $10$ pixels.

Meanwhile, color is another critical important information for human perception.
Color constitutes the main characteristics of the spaces circumscribed by edge lines. Moreover, as a basic low-level feature, it can even impact some high-level concepts such as emotions. Thus, in addition to the edges extracted, we are going to extract the compact color representations, which will be detailed in the next subsection.

\subsection{Compact Representation Extraction}
\label{sec:compress}
% HYY. XXXX. Auto trace. Vectorization. Float2uint. RGB points.
Although edge maps are sparse representations of images, coding such maps into compact bit-streams is still not straight-forward. Existing works in feature-based image compression exploit recurrent generative neural networks~\cite{gregor2016towards} or resort to HEVC Screen Content Coding~\cite{chang2019layered,xu2015overview}. However, these approaches do not fully exploit the sparsity of the edge maps, as they are mostly based on pixel-level representation or partitioning, and not designed to trace edges. It results in inefficiency in coding binary maps consisting only of edges of uniform width.

To explore a more effective way to encode the edge maps, in our approach, we propose to trace the edges into vector graphics. We adopt the image tracing tool~\cite{Autotrace} to convert the binary edge image into vectorized representations. The edges are approximated into straight lines and B{\'e}zier curves, following the Scalable Vector Graphics (SVG) syntax. To be specific, we use three kinds of operation markers, namely \textbf{M}ove, \textbf{L}ine and \textbf{C}urve. Operation \textbf{M}$(x,y)$ indicates moving to point $(x,y)$ without drawing a line. \textbf{L}$(x,y)$ refers to drawing a straight line from the last point (either moved to or ended a line or curve) to the target point $(x,y)$. \textbf{C}$(p_a, p_b, p_t)$ denotes the operation to draw a cubic B{\'e}zier curve from the current point to the target point $p_t$, with the intermediate points $p_a$ and $p_b$. As edge maps of natural images are usually smooth, they can be well approximated by the above-mentioned lines and curves, which only takes a small number of parameters. To further squeeze out redundancy in the parameters, we adopt the Prediction for Partial Matching~(PPM)~\cite{cleary1984data} compression scheme to losslessly compress the quantized parameters for the lines and curves into compact bit-streams.

\begin{figure}[t]
  \centering
  \includegraphics[width=0.98\linewidth]{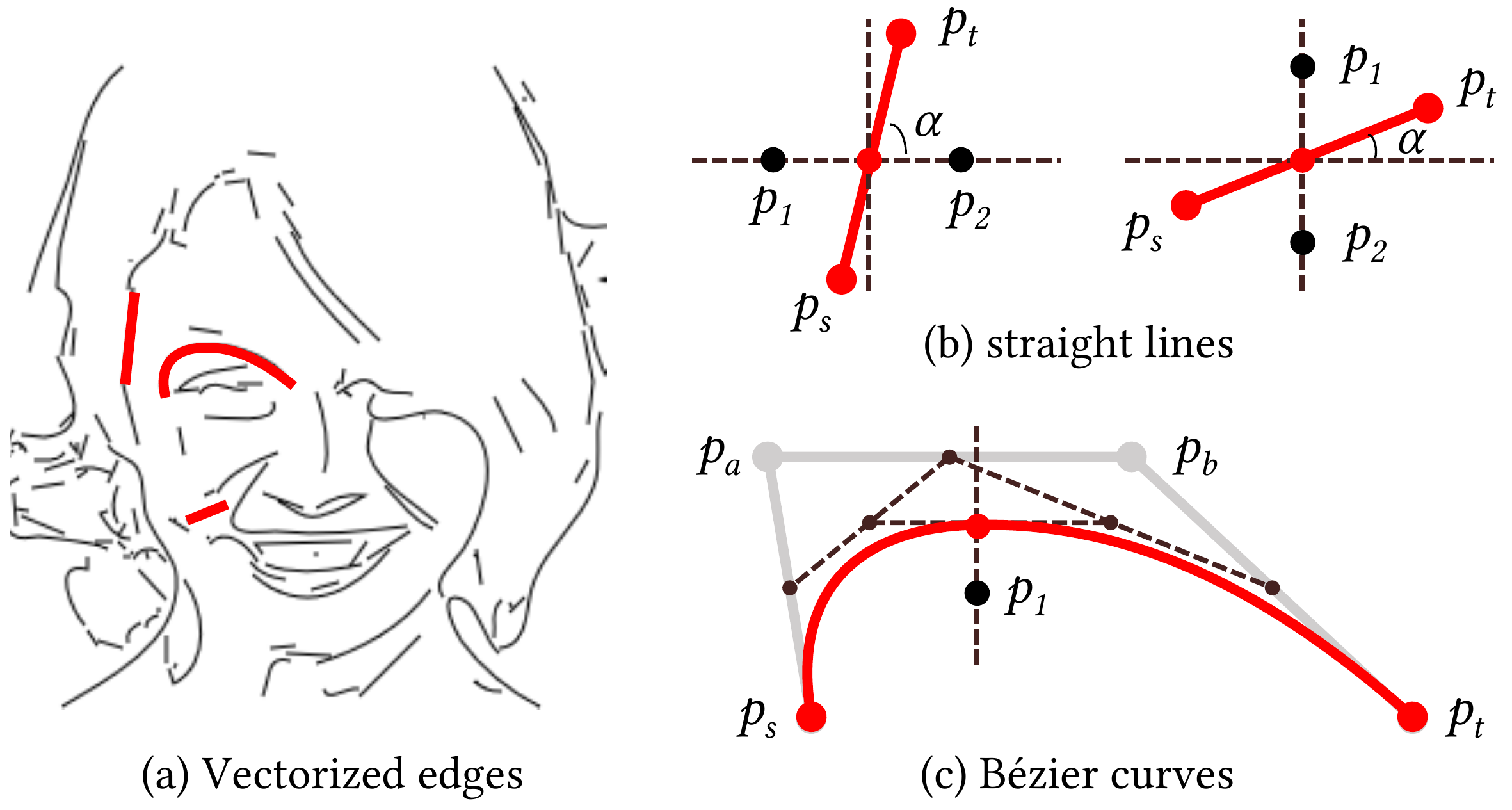}\vspace{-1mm}
  \caption{Illustration of our vectorized structure representation and point samplings for color representation. (a) A vectorized edge map. (b) For straight segments, two points are selected as the reference, according to the slope $\alpha$. (c) For B{\'e}zier curves, one inner point is selected.}
  \label{fig:curve}\vspace{-2mm}
\end{figure}

While edge maps provide much of the information about the structure, the information to restore color representation is lost during the parameterization. To support the scalable coding scheme, we propose to embed pixel-level representation as a second layer in accordance with the encoded structural description. We sparsely sample pixels near the lines and curves. As shown in Fig.~\ref{fig:curve}, for a straight line, we sample two points near the midpoint. The slope of the line is calculated to determine whether the two points are chosen horizontally or vertically. If the line is more close to horizontal ($\alpha < 45^{\circ}$), the two reference points are sampled vertically, and it goes horizontal if $\alpha > 45^{\circ}$. For a B{\'e}zier curve with starting point $p_s$, intermediate points $p_1, p_2$ and target point $p_t$, we first extract the contact point of the curve and the tangent line in parallel with the vector $\overrightarrow{p_sp_t}$. We calculate the slope of the tangent line to determine whether to choose the point vertically or horizontally, just like the straight lines. Additionally, to control the bit-rate and maintain the most informative information from the pixels, we only sample the point at the inner side of the curve, which is expected to have greater gradients and contain more information. The pixel, represented in RGB value, is signaled to the decoder in order as a second layer to provide more fidelity in color. The decoder places the received reference color points following the same rules that the encoder extracts those points, based on the edge maps. Thus, no additional bits are needed to record the positions of the selected pixels.

\begin{figure*}[t]
  \centering
  \includegraphics[width=0.98\linewidth]{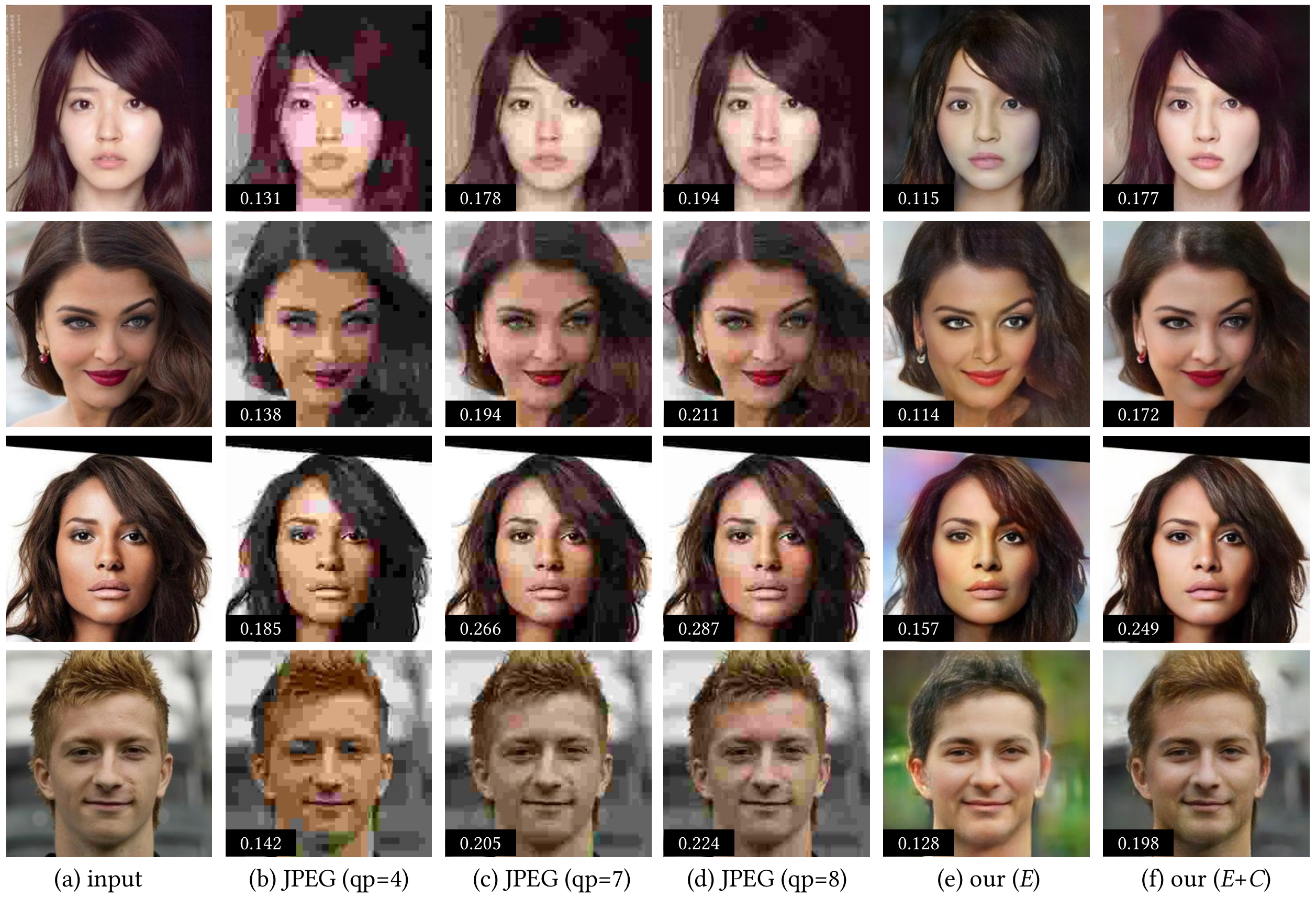}%\vspace{-1mm}
  \caption{Visual comparison with JPEG compression. (a) Input image. (b)-(d) Images compressed by JPEG using quality parameter of $4$, $7$ and $8$, respectively. (e) Our decoded images using the encoded edge representations. (f) Our decoded images using both the encoded edge representation and color representation. For each reconstructed image, its bit-rate (bit per pixel, bpp) is shown in the lower left black box.}\vspace{-3mm}
  \label{fig:comparison}
\end{figure*}

\subsection{Adversarial-based Image Reconstruction}
\label{sec:reconstruct}

Given the proposed compact representation of edges and colors, we are going to recover an image as close as possible to the original image. The main idea is to leverage GAN to learn robust data distribution, which maps our sparse representation back to the original image spaces and benefits both human visual quality and machine visual tasks.

Specifically, we first convert our compact representation back to the image domain by rendering the vector graphic as a normal bitmap $E$.~The sparsely sampled points are rendered as a one-channel image mask $M$ where $1$ means the corresponding pixel is sampled and $0$ vice versa.
And finally, another three-channel image $C$ is provided with the color values of the sampled pixels at the corresponding locations. The remaining unknown pixels are set to $0$.
Through the conversion, we transform our decoding task as a standard machine vision task of image inpainting augmented with extra edge information. $C$ can be regarded to be obtained by the original image $I$ with missing regions indicated by $M$.

Taking use of the advancement of image inpainting research, we design our decoding network as pix2pix~\cite{Isola2017Image}. It contains fully convolutional encoders and decoders, where the low-level information is conveyed to the outputs via skips connections to enforce the structure and color constraints from the inputs. Let our generator and discriminator denoted as $G$ and $D$, respectively. Then $G$ is used to map the input of $E$, $C$ and $M$ to a reconstructed image $I_G=G(E,C,M)$ to approach $I$ in both color and structure senses through a reconstruction loss:
\begin{equation}\label{eq:G_loss}
  \mathcal{L}_{r}=\mathbb{E}\left[\lambda_1\|I_G-I\|+\lambda_2\text{SSIM}(I_G,I)\right],
\end{equation}
where $L_1$ measures the color discrepancy between the reconstructed image and $I$ and SSIM~\cite{wang2004image} emphasizes the structural similarity, weighted by $\lambda_1$ and $\lambda_2$, respectively. In additional to these human-perceptual criteria, we incorporate perceptual loss~\cite{Johnson2016Perceptual} to
enhance the machine-perceptual quality of $I_G$,
\begin{equation}\label{eq:G_loss2}
  \mathcal{L}_{p}=\mathbb{E}\left[\lambda_3\text{PERC}(I_G,I)\right].
\end{equation}
Finally, we use hinge loss~\cite{yu2018free} as our adversarial objective function to learn the data distribution:
\begin{equation}
  \mathcal{L}_{G}=-\mathbb{E}[D(I_{G},E,M)],
\end{equation}
\begin{equation}\label{eq:adv_loss}
\begin{aligned}
  \mathcal{L}_{D}=&~\mathbb{E}[\text{ReLU}(\tau+D(I_{G},E,M))]\\
  +&~\mathbb{E}[\text{ReLU}(\tau-D(I,E,M))],
\end{aligned}
\end{equation}
where $\tau=10$ is a margin parameter.
Here we use channel-wise concatenation to feed multiple inputs into $G$ and $D$.

\noindent\textbf{Reconstruct without RGB}. Note that for some high-level machine vision tasks such as image segmentation and image detection that do not rely on color information much, our framework is scalable to reconstruct images purely from $E$ without $M$ and $C$, which further saves bit-rate.
To be specific, we only need to revise the input channel number of the first layer of $G$ and $D$ such that $G$ receives one-channel input $E$ and $D$ receives four-channel input $(I,E)$ or $(I_G,E)$, respectively.
Beyond that, other settings are the same as our aforementioned reconstruction process with color information.

\begin{figure}[t]
  \centering
  \includegraphics[width=0.98\linewidth]{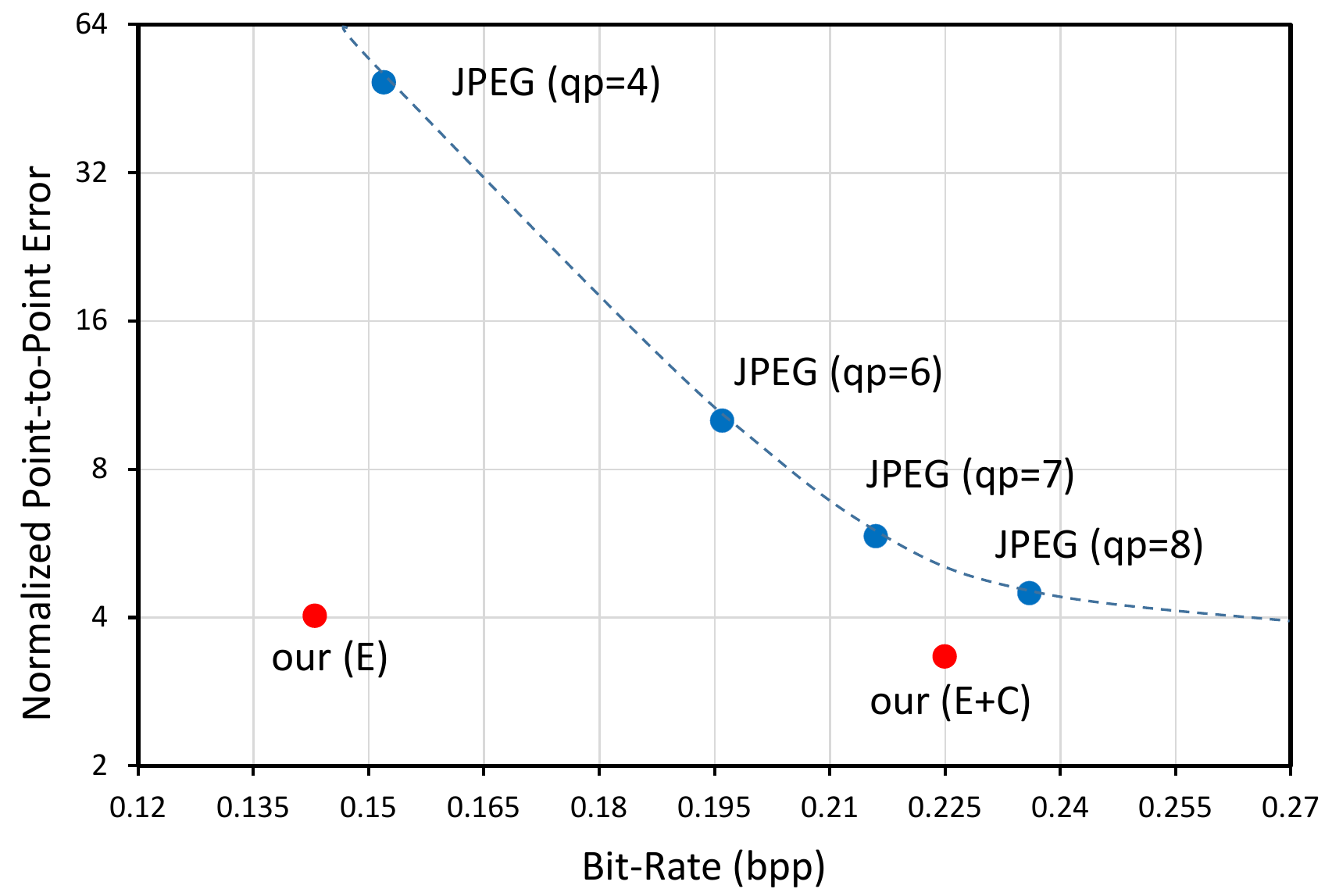}\vspace{-2mm}
  \caption{Illustration of the averaged normalized point-to-point error (NME) on facial landmark detection and bit-rate of JPEG compression and the proposed method.}
  \label{fig:NME}
\end{figure}

\section{Experimental Results}
\label{sec:experiment}

In this section, we present the experimental results of the proposed method for both the task of human vision and machine vision.
We first evaluate our method with respect to human visual quality both qualitatively and quantitatively in Section~\ref{sec:comparison1}.
Then we test our method on the high-level facial landmark detection task in Section~\ref{sec:comparison2}.
We choose the VGGFace2~\cite{cao2018vggface2} dataset for evaluation considering the pervasiveness and importance of facial images in our daily life and industry. We filter the images in VGGFace2 that have small resolution and low quality, and finally use 39,122 images from the training set to train our reconstruction network and 20,665 images from the testing set for performance evaluation. To train our network, we set $\lambda_1=100$, $\lambda_3=1$ and $\tau=10$. $\lambda_2$ is set to $0$ and $1000$ for the human vision evaluation and machine vision evaluation, respectively.

\vspace{-2mm}
\subsection{Human Vision: Visual Quality Evaluation}
\label{sec:comparison1}

\textbf{Qualitative evaluations}.
In Fig.~\ref{fig:comparison}, we present a visual comparison of the proposed method with JPEG compression under different quality parameters~(qp), which are selected to matches the bit-rate of our method for fair comparison. Specifically, for our reconstructed images decoded without color cues, we show the JPEG compression results with $qp=4$; while for our reconstructed images decoded with full color and structure cues, we use $qp=7$ and $qp=8$. It can be observed that JPEG compression yields distinct block artifacts, which greatly decrease visual quality. By comparison, our method produces more natural results.

\noindent\textbf{Quantitative evaluations}. We perform user studies for quantitative evaluations. Besides the four cases shown in Fig.~\ref{fig:comparison}, we randomly select six cases to add up to $10$ cases from the testing data shown to the participants. Each subject is asked to select one from the five results that best matches the original image (\textbf{Fidelity}) and has the best visual quality (\textbf{Aesthetics}). %To ensure the fairness, the orders of five results randomly change every round.
A total of 10 subjects participate in this study and a total of 200 selections are tallied.
The preference ratio is used as the evaluation metric. It is calculated as the ratio of a method selected in all comparisons with this method. As shown in Table~\ref{tb:userstudy}, the proposed structure-color-hybrid method obtains the best average preference ratio of $0.90$ and $0.73$ for both the fidelity and aesthetics, respectively, outperforming JPEG compression under the similar bit-rate. The user study quantitatively verifies the superiority of our method.

\begin{figure}[t]
  \centering
  \includegraphics[width=0.98\linewidth]{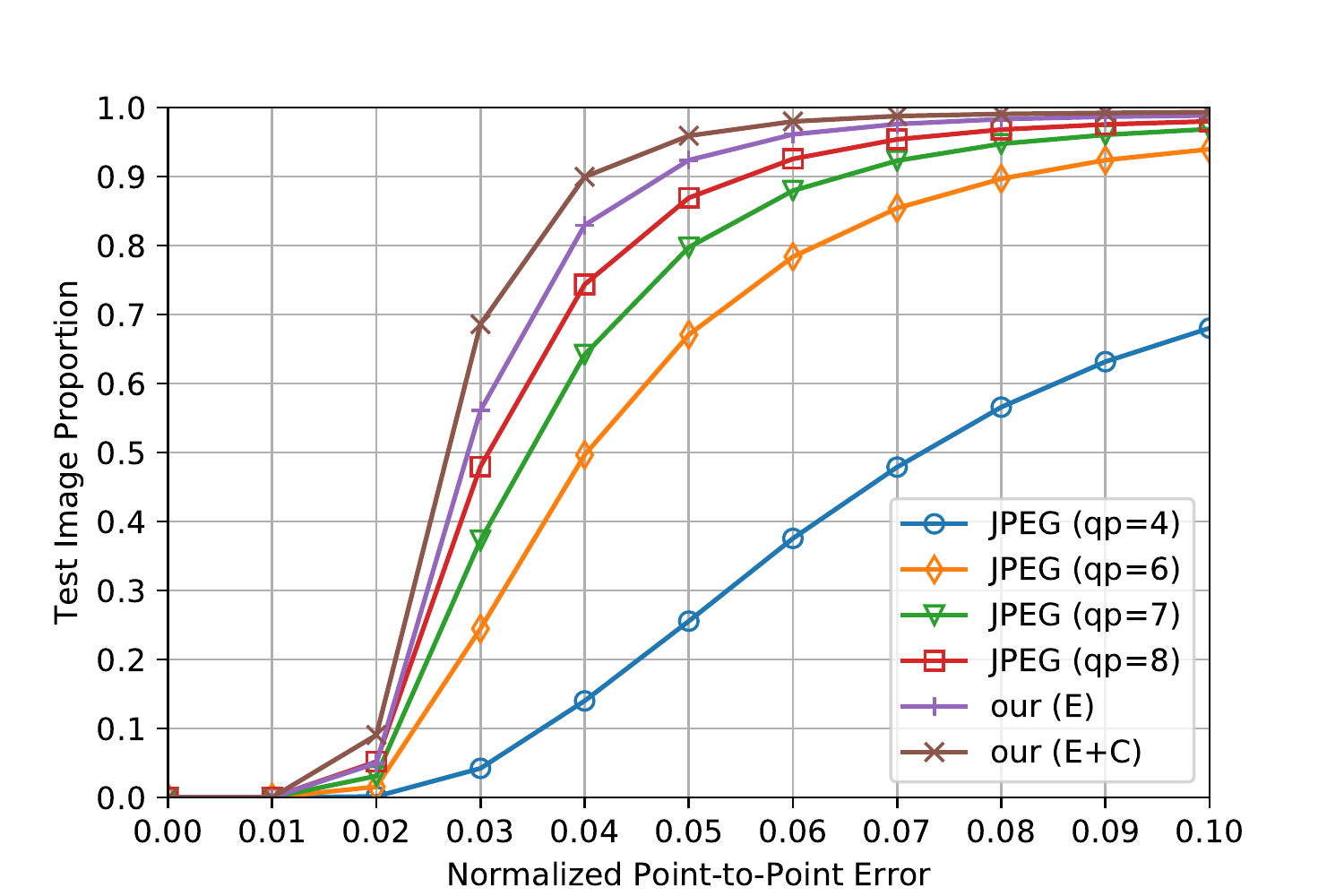}\vspace{-2mm}
  \caption{Cumulative error distribution of JPEG compression and the proposed method on facial landmark detection.}
  \label{fig:CED}
\end{figure}

\begin{table}[t]
    \begin{center}
    \caption{The preference ratio on fidelity and aesthetics of different methods at different bit-rates.}
    \label{tb:userstudy}
    \begin{tabular}{l|ccc}
    \toprule
    Method & Bit-Rate (bpp) & Fidelity & Aesthetics \\
    \midrule
    JPEG ($qp=4$) & 0.152 & 0.00 & 0.00 \\
    our ($E$)  & 0.134 & 0.04 & 0.24 \\
    \midrule
    JPEG ($qp=7$) & 0.214 & 0.02 & 0.01 \\
    JPEG ($qp=8$) & 0.234 & 0.04 & 0.02 \\
    our ($E+C$) & 0.209 & \textbf{0.90} & \textbf{0.73} \\
	\bottomrule
    \end{tabular}
    \end{center}\vspace{-4mm}
\end{table}

\subsection{Machine Vision: Landmark Detection}
\label{sec:comparison2}

The machine vision performance of our method is verified on the high-level facial landmark detection task.
We perform facial landmark detection~\cite{bulat2017far} on the original VGGFace2~\cite{cao2018vggface2} dataset and the reconstructed dataset by JPEG and our method. Detection results on the original data are served as ground truth. We then calculate the normalized point-to-point error (NME)~\cite{cristinacce2006feature} between the detection results on the compressed data and the ground truth.
Fig.~\ref{fig:NME} illustrates the averaged NME and the bit-rate of JPEG compression and our method.
It can be clearly seen that our method achieves much fewer errors at the similar bit-rate compared to JPEG.
Specifically, NME of our method without color cues is only $4.03\%$, which is $44.75\%$ lower than JPEG under $qp=4$. Meanwhile, with color cues, our method achieves merely $3.33\%$ NME, $1.15\%$ lower than JPEG under $qp=8$.
Fig.~\ref{fig:CED} further shows the cumulative error distribution, where more than $90\%$ of the images reconstructed by the proposed method have tiny errors less than $5\%$, showing great robustness.
\vspace{-4mm}
\section{Conclusion and Discussion}
\label{sec:conclusion}
\vspace{-2mm}
In this paper, we present a new image coding framework to facilitate both human vision and machine vision.
The input image is first analyzed and compressed as the compact structure and color representations. Leveraging the advanced generative model in machine vision, we train a network to faithfully reconstruct images from the compact representations. Experimental results demonstrate the superiority of the proposed method in both human visual quality and facial landmark detection. %As a future direction, we still observe a some compression space for color information. One may explore a more smart scalable color point sampling mechanism based on local pixel-level continuity and high-level semantics.
This paper presents the first attempt towards VCM with respective to image coding via scalable feature-based compression. As a future direction, we would like to explore temporal feature modeling for video coding to more pervasively benefit human vision and machine vision.

\bibliographystyle{ieee}
\small
\bibliography{MVC-icme2020}

\end{document}